\documentclass[10pt,twocolumn,letterpaper]{article}

\PassOptionsToPackage{table}{xcolor}

\usepackage[pagenumbers]{cvpr} 

\usepackage{amsmath,amsfonts,bm}

\def\eqref#1{equation~\ref{#1}}

\def\1{\bm{1}}

\def\vh{{\bm{h}}}

\def\vm{{\bm{m}}}

\def\vp{{\bm{p}}}

\def\vt{{\bm{t}}}
\def\vu{{\bm{u}}}

\def\vx{{\bm{x}}}

\def\vz{{\bm{z}}}

\def\mX{{\bm{X}}}

\DeclareMathAlphabet{\mathsfit}{\encodingdefault}{\sfdefault}{m}{sl}
\SetMathAlphabet{\mathsfit}{bold}{\encodingdefault}{\sfdefault}{bx}{n}

\def\gL{{\mathcal{L}}}

\def\sM{{\mathbb{M}}}

\def\sR{{\mathbb{R}}}

\def\sV{{\mathbb{V}}}

\usepackage{cuted}          
\usepackage{tikz}
\usepackage{pgfplots}
\pgfplotsset{compat=1.18}
\usetikzlibrary{arrows.meta}
\usetikzlibrary{calc}
\usetikzlibrary{positioning}
\usepackage{pifont}
\usepackage{multirow}
\usepackage{makecell}
\usepackage{comment}        
\usepackage{microtype}      

\newlength{\titlegap}
\iftoggle{cvprfinal}{\iftoggle{cvprpagenumbers}{\setlength{\titlegap}{-12pt}}{}}{}

\providecommand{\cmidruleheight}{\dimexpr(\aboverulesep+\cmidrulewidth+\belowrulesep)/2\relax}

\pgfdeclareplotmark{disc}{%
  \pgfpathcircle{\pgfpoint{0pt}{0pt}}{\pgfplotmarksize}%
  \pgfusepathqfillstroke%
}

\definecolor{full_color}{HTML}{DD8800}
\definecolor{mae_color}{HTML}{000000}
\definecolor{half_color}{HTML}{EE0000}

\definecolor{green_weak_color}{HTML}{0000FF}
\definecolor{none_medium_color}{HTML}{00CC00}
\definecolor{none_strong_color}{HTML}{CC33FF}

\colorlet{full_color_strong}{full_color!80!black}
\colorlet{half_color_strong}{half_color!80!white}

\definecolor{good_color}{HTML}{00BB00}
\definecolor{bad_color}{HTML}{FF2200}

\pgfplotsset{
    lejepamarker/.style={color=gray, mark=diamond*, mark size=2.5pt},
    simdinov2marker/.style={color=gray, mark=oplus, mark size=2.5pt},
    bootlegmarker/.style={color=gray, mark=square, mark size=2.5pt},
    colormaemarker/.style={color=gray, mark=o, mark size=2.5pt, dashed, mark options={solid}},
    ogmaemarker/.style={color=gray, mark=asterisk, mark size=3.5pt},
    full_strong/.style={color=full_color_strong, dashed, mark=*,
    mark options={solid, fill=full_color_strong, draw=black, line width=0.6pt}},
    half_strong/.style={color=half_color_strong, dashed, mark=triangle*,
    mark options={solid, fill=half_color_strong, draw=black, line width=0.6pt, mark size=2.5pt}},
    green_weak/.style={color=green_weak_color, solid, mark=*,
    mark options={solid, fill=green_weak_color}},
    none_medium/.style={color=none_medium_color, solid, mark=square*,
    mark options={solid, fill=none_medium_color}},
    none_strong/.style={color=none_strong_color, solid, mark=triangle*,
    mark options={solid, fill=none_strong_color}},
    fullmarker/.style={color=full_color, mark=*, mark options={solid}},
    halfmarker/.style={color=half_color, mark=triangle*, mark options={solid, mark size=2.5pt}},
    ourmaemarker/.style={color=mae_color, mark=square*, mark options={solid}},
}

\DeclareRobustCommand{\plotmarker}[2][1pt]{%
  \tikz[baseline=-0.6ex]\draw[line width=#1, /pgfplots/#2]
    plot coordinates {(0,0)};}

\pgfplotsset{
    mylegend/.style={
        hide axis, scale only axis, width=1pt, height=1pt,
        xmin=0, xmax=1, ymin=0, ymax=1,
        legend style={
            at={(0.5,0.5)}, anchor=center, draw=none,
            /tikz/every even column/.append style={column sep=6pt},
            nodes={font=\scriptsize},
        },
        mark size=2pt, line width=1.2pt,
    }
}

\newlength{\subplotwidth}
\newlength{\subplotgap}
\newcommand{\cls}{\texttt{CLS}}

\newcommand{\hider}[1]{}

\definecolor{cvprblue}{rgb}{0.21,0.49,0.74}
\usepackage[pagebackref,breaklinks,colorlinks,allcolors=cvprblue]{hyperref}

\def\confName{CVPR}
\def\confYear{2027}

\newcommand{\papertitle}{Masked Swingers: Harnessing Data Augmentation to\\Advance Autoencoders for Self-Supervised Learning}
\title{\vspace{\titlegap}\papertitle\vspace{\titlegap}}
\renewcommand{\thetitle}{\papertitle}   

\author{
Anthony Fuller$^{1}$\quad Scott C. Lowe$^{1}$\quad Daniel G. Kyrollos$^{2}$ \\
Graham W. Taylor$^{1,3\bigstar}$\quad Evan Shelhamer$^{1,4\bigstar}$\quad James R. Green$^{5\bigstar}$ \\
\normalsize $^{1}$Vector Institute\quad $^{2}$University of Ottawa\quad
$^{3}$University of Guelph\quad $^{4}$University of British Columbia\quad $^{5}$Carleton University \\
\normalsize $^{\bigstar}$Joint last author \\[\titlegap]
}

\begin{document}
\maketitle
\begin{figure*}[t]
\centering
\resizebox{0.7\textwidth}{!}{\input{figures/schematic/figure}}
\caption{
Our \textbf{Masked Swingers} exchanges \textcolor{half_color}{\textbf{\texttt{CLS} tokens}} between different views of the same image prior to reconstruction.
We make different views through independent data augmentation and masking.
Together, these two design choices encourages the encoder to summarize its input to help reconstruct an arbitrary view so that it can be transferred efficiently and broadly.
}
\label{fig:schematic}
\end{figure*}

\begin{abstract}
Self-supervised learning (SSL) trains on inputs alone,
allowing us to make models that are more useful in more settings.
The autoencoder SSL framework learns by reconstructing its own input after information loss through a bottleneck or noise injection.
Masked autoencoders (MAE) are the most successful instantiation of this framework: they encode a random subset of patches, then decode the masked-out patches.
In this work, we introduce key modifications to improve MAEs.
Our method augments an image in two different ways, then masks and encodes each view separately.
It then exchanges the global representations (\cls{} tokens) between views before decoding the masked patches.
By design, our \textbf{Masked Swingers} encourages learning a view-agnostic summary of the image for efficient transfer.
We perform extensive experiments, and find Masked Swingers outperforms the official MAE checkpoint by +10.7\% averaged over 10 frozen-feature tasks at equal pre-training compute.
Gains are highest on fine-grained tasks, e.g., instance retrieval.
Swingers also reduces error by 64\% relative to MAE on three new state-probing datasets for world modeling, while keeping the fine-tunability of MAE.
Post-training by swinging nearest neighbors gains again, e.g., by +5\% on IN-1K $k$NN, opening the door to learning invariances via autoencoding.
\end{abstract}

\section{Introduction: Autoencoding and Views}

Autoencoding~\cite{cottrell1987learning, ballard1987modular} is an old, yet still popular self-supervised learning (SSL) framework.
It consists of an encoder, which receives an input and hands a hidden state to a decoder, which predicts the input---and to learn \emph{useful} representations, it puts an obstacle somewhere in the way, e.g., by masking encoder-input portions~\cite{vincent2008extracting, pathak2016context, he2022masked} or sending the hidden state through a low-dimensional, ``information bottleneck'' \cite{tishby2000information} to learn a summary representation~\cite{bourlard1988auto, hinton2006reducing, van2017neural}.

Joint embedding~\cite{becker1992self, bromley1993signature} is another popular SSL framework.
It separately encodes two views of a sample and encourages their representations to be similar---and to learn \emph{useful} representations, it controls the mutual information between the views, often through data augmentation~\cite{chen2020simple, he2020momentum}.
Because the loss is computed on a single vector per view, the encoder learns to summarize its input.
This insight, that two views of a sample should be represented similarly, underpins many SSL algorithms~\cite{chen2021empirical, bardes2021vicreg}, yet no autoencoding method directly encourages learning view-agnostic representations.

We thus introduce \textbf{Masked Swingers}, which harnesses data augmentation to improve the representations learned from autoencoding.
Our Masked Swingers algorithm augments and masks a sample in two different ways and separately encodes each view.
It then exchanges the encoded global ``\cls{}'' tokens between views---exchanging messages between them---and separately decodes the originally masked portions of each view.
This encourages the encoder to summarize its input to help reconstruct an arbitrary view through a bottleneck.
Because our method augments and masks views independently, it encourages this summary to be useful regardless of the view-specific transformations.

Masked Autoencoders (MAEs)~\cite{he2022masked} offer several advantages.
They have proven effective across data distributions and modalities, including natural imagery~\cite{singh2023effectiveness}, satellite imagery~\cite{cong2022satmae}, 1D time series~\cite{li2023ti}, and 3D point clouds~\cite{pang2022masked}.
They have low device-memory requirements: only one sample at a time must fit in memory for training.
This is because, unlike most joint-embedding methods, autoencoders need no cross-sample negatives \cite{chen2020simple} or statistics \cite{ balestriero2025lejepa} to avoid collapse, and their Vision Transformer backbone (ViT)~\cite{dosovitskiy2021an} does not use batch normalization~\cite{pmlr-v37-ioffe15}.
MAE pre-training also tends to be robust because its targets are stationary/grounded, unlike self-distillation methods~\cite{assran2023self}.
MAE is thus frequently used to pre-train ViTs across settings, so we aim to improve it.

We summarize our contributions and results.
\begin{enumerate*}[label=\textbf{(\roman*)}]
\item We propose a novel autoencoding SSL algorithm called Masked Swingers, which harnesses data augmentation to promote a view-agnostic image summary for efficient transfer.
\item We extensively show our Swingers outperforms MAE across 67 datasets when used as a global, frozen feature extractor---gains are highest on fine-grained tasks, e.g., instance retrieval (+8\%), animal re-identification (+5\%), and species classification (+5\%).
\item We study mask rate and model size, showing that Swingers can benefit from lower mask rates than MAE and that our method scales well across Tiny/Small/Base/Large ViTs.
\item We show initial results in world modeling---Swingers excels at state prediction, reducing relative error by 64\%.
\item We can exchange other tokens, besides the \cls{}, to still perform well.
\item We propose a simple post-training method that exchanges the encoded \cls{} tokens between distinct, yet similar images (found via nearest neighbor over IN-1K train) to gain again, e.g., increases on IN-1K \cite{russakovsky2015imagenet} by +5.3\% and on iNaturalist \cite{van2021benchmarking} by +6.0\%.
\end{enumerate*}

\section{Background: Masked Autoencoders}
\label{sec:background}

We review details of the MAE algorithm because our Masked Swingers inherits much of its design.
 
\textbf{Patchifying.}
An image $\vx {\in} \sR^{H \times W \times C}$ is divided into $N {=} HW/P^2$ non-overlapping patches of resolution $P \times P$ and flattened into a sequence $\mX {=} \left[ \vx_1, \dots, \vx_N \right]$, $\vx_i {\in} \sR^{P^2 C}$.
A linear layer maps these patches to the token dimension $d$, which is the encoder ``width'', and position embeddings $\vp_i {\in} \sR^{d}$ are added to make patch tokens $\vt_i {=} \mathrm{Linear}(\vx_i) {+} \vp_i$.

\textbf{Masking.}
A random subset $\sM$ of $\rho N$ patch indices is dropped, leaving the \emph{visible} patches $\sV$, where $\rho$ is the masking ratio.
MAE samples masks uniformly; ColorMAE~\cite{hinojosa2024colormae} thresholds uniform noise passed through a 2D filter, giving more contiguous visible and masked regions.
``Red'' color masking yields large contiguous masks and ``green'' masking yields smaller regions.
Because of spatial autocorrelation, for the same mask rate, red and green color masking result in less mutual information between independently sampled masks of the same image, relative to uniform masking.

\textbf{Encoding.}
A learnable \cls{} token $\vt_{\cls{}}$ is prepended to the visible tokens.
Crucially, the encoder $f_\theta$ is applied only to these tokens,
$\left[ \vz_{\cls{}}, \left\{ \vz_i \right\}_{i {\in} \sV} \right]
{=} f_\theta \left( \left[ \vt_{\cls{}}, \left\{ \vt_i \right\}_{i {\in} \sV} \right] \right),
 \vz_i {\in} \sR^{d}$.

\textbf{Decoding.}
The decoder operates on all $N$ patch positions plus the \cls{} token.
Its inputs $\vh_i {\in} \sR^{d'}$ are $\vh_i {=} \mathrm{Linear}(\vz_i) {+} \vp'_i$ for visible patches $i {\in} \sV$, projecting to the (narrower) decoder width $d'$, and $\vh_i {=} \vm {+} \vp'_i$ for masked patches $i {\in} \sM$, where $\vm {\in} \sR^{d'}$ is a learnable mask token that is shared across \emph{all} masked positions and $\vp'_i$ are decoder position embeddings.
The \cls{} token uses the same linear layer, $\vh_{\cls{}} {=} \mathrm{Linear}(\vz_{\cls{}})$.
Transformer blocks, $g_\phi$, map these to latents,
$\left[ \vu_{\cls{}}, \vu_1, \dots, \vu_N \right]
{=} g_\phi \left( \left[ \vh_{\cls{}}, \vh_1, \dots, \vh_N \right] \right)$.
We discard $\vu_{\cls{}}$ and linearly map each patch latent to pixel predictions, $\hat{\vx}_i {=} \mathrm{Linear}(\vu_i) {\in} \sR^{P^2 C}$.
 
\textbf{Objective.}
The loss,
$\gL_{\mathrm{MAE}}(\theta, \phi) {=} \frac{1}{|\sM|} \sum_{i {\in} \sM} \lVert \hat{\vx}_i {-} \tilde{\vx}_i \rVert_2^2$,
is the mean squared error on masked positions only.
Targets are pixels, $\tilde{\vx}_i {=} (\vx_i {-} \mu_i) / \sigma_i$, normalized by the mean $\mu_i$ and standard deviation $\sigma_i$ of patch~$i$.

\section{Method: Masked Swingers}
\label{sec:method}

Masked Swingers changes one thing about MAE: the \cls{} token the decoder receives is from the encoder applied to a \emph{different} view of the same sample, created via applying independent masking and data augmentation (Fig. \ref{fig:schematic}).

\textbf{Two views.}
We augment each image twice, into $\vx^{1}$ and $\vx^{2}$, which we patchify and mask independently, giving visible sets $\sV^{1}, \sV^{2}$ and masked sets $\sM^{1}, \sM^{2}$.
We encode the two views separately by the same encoder $f_\theta$, producing $\vz^{1}_{\cls{}}, \vz^{2}_{\cls{}}$ and patch tokens $\{\vz^{1}_i\}_{i {\in} \sV^{1}}$, $\{\vz^{2}_i\}_{i {\in} \sV^{2}}$.

\textbf{Swinging.}
We decode the input exactly as in MAE, from each view's own patch tokens and mask tokens, with one key change: we trade the \cls{} tokens between views,
$\vh^{1}_{\cls{}} {=} \mathrm{Linear}(\vz^{2}_{\cls{}})$ and $\vh^{2}_{\cls{}} {=} \mathrm{Linear}(\vz^{1}_{\cls{}})$.
Everything downstream is unchanged from MAE: we map each sequence to latents with $g_\phi$, we discard the \cls{}, and we project the remaining tokens back to pixels.

\textbf{Objective.}
We compute the loss as the MAE reconstruction error of Sec.~\ref{sec:background}, averaged over both views,
$\gL_{\mathrm{Swingers}}(\theta, \phi) {=} \frac{1}{2} \sum_{v {\in} \{1,2\}} \frac{1}{|\sM^{v}|} \sum_{i {\in} \sM^{v}} \lVert \hat{\vx}^{v}_i - \tilde{\vx}^{v}_i \rVert_2^2$.

Since the masked patches of $\vx^{1}$ must be inferred from its own visible patches and the \cls{} token of $\vx^{2}$ (and symmetrically), the encoder must compress each view into a summary that stays useful under a different augmentation and a different mask---promoting a view-agnostic summary.
We call this algorithm ``\textbf{Full-Swingers}'', to distinguish it from the Masked-Swinger variant that we introduce next.

\textbf{Half-Swingers.}
We propose a simple variant that uses the \cls{} token from one view to decode both views (this is equivalent to exchanging \cls{} tokens on 50\% of images and keeping the same-view \cls{} tokens for the other 50\%),
$\vh^{1}_{\cls{}} {=} \mathrm{Linear}(\vz^{1}_{\cls{}})$ and $\vh^{2}_{\cls{}} {=} \mathrm{Linear}(\vz^{1}_{\cls{}})$.
Our ``Half-Swingers'' trains the encoder to simultaneously
\begin{enumerate*}[label=\textbf{(\roman*)}]
\item summarize its own view to help decode arbitrary views and
\item infer the masked patches of its own view.
\end{enumerate*}
The encoder is not aware if its \cls{} helps reconstruct its own view or another view, and the decoder is not aware from which view its \cls{} originates---so the encoder may compromise between ``compressing the visible'' and ``inferring the missing''.

\section{Experiments: Controls and Context}

In Sec.~\ref{sec:pre_train_aug_color_epochs}, we run controlled experiments to compare our Masked Swingers to MAE, avoiding confounds that could otherwise account for performance gaps.
We report the performance of models pre-trained using other frameworks for reference and not for direct comparison.
Since self-supervised representations should be general, we evaluate on IN-1K~\cite{russakovsky2015imagenet}, the gold-standard, alongside other datasets that check model performance for other uses, e.g. instance retrieval~\cite{kordopatis2025ilias, oxford_paris, oh2016deep} and fine-grained tasks~\cite{van2021benchmarking, Picek_2025_CVPR}.
We also fine-tune on IN-1K, OmniBenchmark \cite{zhang2022benchmarking}, and ADE20K \cite{zhou2017scene} to check if Swingers retains MAE's fine-tunability.
In Sec.~\ref{sec:swing_rate}, we study the swing rate (how often we exchange \cls{} tokens) alongside augmentation, and in Sec.~\ref{sec:mask_rate_and_model_size}, we study the effects of masking rate and model size on both Swinger variants because these are two important design axes.
In Sec.~\ref{sec:wm}, we explore if Masked Swingers can help with world modeling, as a new and exciting use of pre-trained vision encoders.
We do this by probing the features extracted by models for ``state'' information on three new datasets~\cite{motionjepa}.
Then in Sec.~\ref{sec:analysis_ablations_alternatives}, we provide:
\begin{enumerate*}[label=\textbf{(\roman*)}]
\item analysis, by looking at data augmentation's interaction with pre-training length;
\item ablations, by giving both encoders the same augmentation to confirm our motivating hypothesis; and 
\item alternatives, by exchanging tokens besides the \cls{} or adding tokens.
\end{enumerate*}
In Sec.~\ref{sec:nn}, we swing distinct, yet similar instances (their nearest neighbors) during a short post-training stage which further boosts $k$NN substantially.
Finally, in Sec.~\ref{sec:infonce}, we show that Swingers achieves relative view invariance---achieving lower InfoNCE loss \cite{infonce} without directly optimizing for it.

\begin{figure*}[t!]
\centering

\begin{subfigure}[t]{\subplotwidth}
    \input{figures/main_eval/imagenet_knn}
\end{subfigure}%
\hspace{\subplotgap}%
\begin{subfigure}[t]{\subplotwidth}
    \input{figures/main_eval/imagenet_21k_knn}
\end{subfigure}%
\hspace{\subplotgap}%
\begin{subfigure}[t]{\subplotwidth}
    \input{figures/main_eval/instance_retrieval}
\end{subfigure}%
\hspace{\subplotgap}%
\begin{subfigure}[t]{\subplotwidth}
    \input{figures/main_eval/animal_reidentification}
\end{subfigure}%
\hspace{\subplotgap}%
\begin{subfigure}[t]{\subplotwidth}
    \input{figures/main_eval/inat}
\end{subfigure}
\\[-2ex]
\begin{subfigure}[t]{\subplotwidth}
    \input{figures/main_eval/vtab_natural}
\end{subfigure}%
\hspace{\subplotgap}%
\begin{subfigure}[t]{\subplotwidth}
    \input{figures/main_eval/vtab_specialized}
\end{subfigure}%
\hspace{\subplotgap}%
\begin{subfigure}[t]{\subplotwidth}
    \input{figures/main_eval/vtab_structured}
\end{subfigure}%
\hspace{\subplotgap}%
\begin{subfigure}[t]{\subplotwidth}
    \input{figures/main_eval/characters}
\end{subfigure}%
\hspace{\subplotgap}%
\begin{subfigure}[t]{\subplotwidth}
    \input{figures/main_eval/fungi}
\end{subfigure}
\\[-3ex]

\input{figures/main_eval/legend}
\\[-2ex]
\caption{\textbf{Our \textcolor{half_color}{Half-Swingers} extracts better features than MAE} on all 10 tasks and 3 pre-training lengths.
Our \textcolor{full_color}{\textbf{Full-Swingers}} beats MAE on instance retrieval, animal re-identification, mini-VTAB specialized, and Omniglot, but trails it on natural image classification.
In our controlled experiments (\plotmarker{fullmarker} \plotmarker{halfmarker} \plotmarker{ourmaemarker}), we show one recipe (data augmentation and color masking strengths) per SSL algorithm, pre-trained for 100, 200, and 800 epochs; it is the best of 9 recipes by average accuracy over all 10 tasks at 800 epochs.
Intended for reference only, we also run evaluations on public/official pre-trained checkpoints (\plotmarker{ogmaemarker} \plotmarker{colormaemarker} \plotmarker{lejepamarker} \plotmarker{simdinov2marker} \plotmarker{bootlegmarker}).
}
\label{fig:main_eval}
\end{figure*}

\subsection{Pre-training: Varying Augmentation, Color Masking, and Epochs}
\label{sec:pre_train_aug_color_epochs}

\textbf{Pre-training Setup: Varied Settings.}
We compare three autoencoding SSL algorithms: the MAE/ColorMAE baseline and our two Masked-Swinger variants, Full-Swingers and Half-Swingers.
For each algorithm, we pre-train 27 models spanning the cross product of pre-training length \{100, 200, 800 epochs\}, data augmentation strength \{weak, medium, strong\}, and color masking \{none, green, red\}.
Our weak, medium, and strong settings apply RandAug~\cite{cubuk2020randaugment} with progressively more operations at higher magnitudes.
We vary augmentation and color masking because they control the shared information between views, which is important for Swingers.
We also vary these knobs for the MAE baseline to equally tune it, ensuring experimental fairness.

\textbf{Pre-training Setup: Fixed Settings.}
We use the ViT-B/16 architecture~\cite{dosovitskiy2021an}, which is the most common ViT size, and study other model sizes in Sec.~\ref{sec:mask_rate_and_model_size}.
We pre-train on ImageNet-1K at $224 \times 224$ resolution.
Since Swingers encodes two views per image, we train the MAE baseline with a repeat augmentation~\cite{berman2019multigrain, hoffer2020augment} of 2, i.e., each image appears twice per batch under independent augmentations and masks.
All three algorithms thus process the same number of views per step, so we hold the batch size, learning rate, and total pre-training compute fixed across them; the algorithm implementations differ by only a few lines of code.
For all other settings, including optimization hyperparameters, we use MAE's defaults, since Swingers shares MAE's reconstruction objective and should have similar optimal values.

\textbf{Other models as references.}
To contextualize our controlled experiments, we also evaluate publicly released models, which differ from ours in their pre-training algorithms or recipes and are not meant for direct comparison.
First, we evaluate the official MAE and ColorMAE checkpoints.
These models differ from our controlled MAE/ColorMAE baselines in learning rate and batch size, and they pre-train without repeat augmentation.
Second, we evaluate three strong and recent \emph{non-autoencoding} models: LeJEPA~\cite{balestriero2025lejepa}, SimDINOv2~\cite{wu2025simplifying}, and Bootleg~\cite{lowe2026self}.
Like ours, these models use ViT-B/16 and pre-train on ImageNet-1K at $224 \times 224$ resolution.
LeJEPA has no public ViT-B/16 checkpoint, so we use a public reproduction (supplementary Sec.~\ref{sec:appendix_references}).

\textbf{Evaluation Setup: Frozen Features.}
We evaluate frozen features on 10 tasks spanning 67 datasets:
\begin{enumerate*}[label=\textbf{(\roman*)}]
\item $k$NN on ImageNet-1K~\cite{russakovsky2015imagenet};
\item $k$NN on ImageNet-21K~\cite{ridnik2021imagenetk} with limited labels, using 10 training images from each of its 10,450 classes (104.5K images) and all 522.5K validation images;
\item $k$NN on mini-VTAB~\cite{fuller2026selfsoupervision}, which is a smaller version of VTAB~\cite{zhai2019large} with 21 tasks in total, split into its \textbf{(a)} natural, \textbf{(b)} specialized, and \textbf{(c)} structured subsets;
\item instance retrieval on Revisited Oxford and Paris~\cite{oxford_paris} (ROxford5k and RParis6k, medium protocol), Stanford Online Products~\cite{oh2016deep}, and the ILIAS core set~\cite{kordopatis2025ilias}, without YFCC100M distractors~\cite{thomee2016yfcc100m};
\item animal re-identification on WildlifeReID-10k~\cite{adam2025wildlifereid}, which covers 10.7K individuals from 37 datasets, using the standard open-set split with a single pooled gallery;
\item handwritten character recognition on Omniglot's 20-way one-shot benchmark~\cite{lake2015human};
\item $k$NN on iNaturalist 2021-mini~\cite{van2021benchmarking}, with 10K species, 50 training images each (500K total), and all 100K validation images; and
\item $k$NN on the long-tailed closed-set split of FungiTastic~\cite{Picek_2025_CVPR}, with 2.8K fungal species, 433.7K training images, and 89.7K validation images.
\end{enumerate*}
We report mAP for instance retrieval and animal re-identification, and top-1 accuracy for all other tasks.
For $k$NN, we tune the number of neighbors and the softmax temperature per model and dataset.
We evaluate both the \cls{} token and the mean-pooled patch tokens, and report the better of the two on each dataset (see supplementary Sec.~\ref{sec:appendix_frozen}).
For each algorithm, we plot results obtained by a single augmentation and color-mask recipe (chosen based on the average of all 10 tasks at 800 epochs).

\textbf{Results: Masked Swingers learns better global, frozen features than MAE on all 10 tasks (Fig.~\ref{fig:main_eval}).}
On IN-1K $k$NN, Half-Swingers outperforms MAE by +6.4/+8.6/+4.2\% at 100/200/800 pre-training epochs.
Gains are relatively larger on the harder 10-shot IN-21K task, which has more classes and fewer examples per class: Half-Swingers boosts accuracy +1.7\%/+2.6\%/+2.2\% over MAE.
Swingers' gains tend to be larger still on the remaining 8 tasks.
On Omniglot at 800 epochs, for example, Full-Swingers and Half-Swingers reach 83\% and 77\% accuracy versus 48\% for MAE.
These results use a single setting per algorithm (Fig.~\ref{fig:main_eval}); Half-Swingers also outperforms MAE when averaged over all 9 augmentation and color-masking settings, by +5.6/+4.8/+0.5\% on IN-1K $k$NN.
The trend with pre-training length depends on the task.
Versus MAE's official 1600-epoch pre-trained checkpoint (\plotmarker{ogmaemarker}) our Half-Swingers wins by +10.7\% averaged over the 10 tasks, which spans 67 datasets.
Please see supplementary Sec.~\ref{sec:appendix_all_results} for all results.

\begin{table}[t]
\centering
\scriptsize
\setlength{\tabcolsep}{2pt}
\providecommand{\fttbd}{\textcolor{red}{xx.x}}
\providecommand{\lpmae}{MAE/ColorMAE}
\providecommand{\lphalf}{\textcolor{half_color}{\textbf{Half-Swingers}}}
\providecommand{\lpfull}{\textcolor{full_color}{\textbf{Full-Swingers}}}
\providecommand{\lphalfbg}{\rowcolor{half_color!10}}
\providecommand{\lpfullbg}{\rowcolor{full_color!10}}
\begin{minipage}[c]{0.51\linewidth}
    \caption{
    \textbf{Half-Swingers performs similarly to MAE on IN-1K linear probing.}
    Backbones chosen with winning recipes from our 10-task frozen-feature evals (Fig. \ref{fig:main_eval}).
    }
    \label{tab:in1k_lp}
\end{minipage}%
\hfill
\begin{minipage}[c]{0.47\linewidth}
    \vspace{-10pt}
    \centering
    \begin{tabular}{lcc}
         & \multicolumn{2}{c}{ImageNet-1K} \\
        \cmidrule(lr){2-3}
        Model & 100 ep. & 800 ep. \\
        \toprule
        \lpmae & 51.4 & \textbf{65.0} \\
        \lphalfbg \lphalf & \textbf{53.9} & 63.9 \\
        \lpfullbg \lpfull & 40.5 & 54.2 \\
        \bottomrule
    \end{tabular}
\end{minipage}
\end{table}

\begin{table}[t]
\centering
\scriptsize
\setlength{\tabcolsep}{4.6pt}
\providecommand{\fttbd}{\textcolor{red}{xx.x}}
\providecommand{\lpmae}{MAE/ColorMAE}
\providecommand{\lphalf}{\textcolor{half_color}{\textbf{Half-Swingers}}}
\providecommand{\lpfull}{\textcolor{full_color}{\textbf{Full-Swingers}}}
\providecommand{\lphalfbg}{\rowcolor{half_color!10}}
\providecommand{\lpfullbg}{\rowcolor{full_color!10}}
\caption{
\textbf{Our Masked Swingers are as fine-tunable as MAE.}
For OmniBenchmark and ADE20K, we select the pre-trained checkpoints which yielded the best performance when fine-tuning on IN-1K and fine-tune only these checkpoints on OmniBenchmark and ADE20K.
For IN-1K and OmniBenchmark, we report top-1 acc.
For ADE20K, we use mIoU.
For OmniBenchmark, we average over all 21 datasets.
100 and 800 ep. refers to pre-training epochs.
}
\label{tab:main_ft}
\begin{tabular}{lcccccc}
     & \multicolumn{2}{c}{ImageNet-1K} & \multicolumn{2}{c}{OmniBenchmark} & \multicolumn{2}{c}{ADE20K} \\
    \cmidrule(lr){2-3}\cmidrule(lr){4-5}\cmidrule(lr){6-7}
    Model & 100 ep. & 800 ep. & 100 ep. & 800 ep. & 100 ep. & 800 ep. \\
    \toprule
    \lpmae & 81.5 & 83.0 & 41.2 & 44.9 & \textbf{44.6} & 46.5 \\
    \lphalfbg \lphalf & 81.6 & \textbf{83.1} & 42.9 & \textbf{45.2} & 44.2 & \textbf{46.8} \\
    \lpfullbg \lpfull & \textbf{81.9} & \textbf{83.1} & \textbf{43.1} & 44.6 & 43.7 & 45.3 \\
    \bottomrule
\end{tabular}
\end{table}

\begin{figure*}[t!]
\centering

\begin{subfigure}[t]{\subplotwidth}
    \input{figures/swing_rate/imagenet_knn}
\end{subfigure}%
\hspace{\subplotgap}%
\begin{subfigure}[t]{\subplotwidth}
    \input{figures/swing_rate/imagenet_21k_knn}
\end{subfigure}%
\hspace{\subplotgap}%
\begin{subfigure}[t]{\subplotwidth}
    \input{figures/swing_rate/instance_retrieval}
\end{subfigure}%
\hspace{\subplotgap}%
\begin{subfigure}[t]{\subplotwidth}
    \input{figures/swing_rate/animal_reidentification}
\end{subfigure}%
\hspace{\subplotgap}%
\begin{subfigure}[t]{\subplotwidth}
    \input{figures/swing_rate/inat}
\end{subfigure}
\\[-2.5ex]
\begin{subfigure}[t]{\subplotwidth}
    \input{figures/swing_rate/vtab_natural}
\end{subfigure}%
\hspace{\subplotgap}%
\begin{subfigure}[t]{\subplotwidth}
    \input{figures/swing_rate/vtab_specialized}
\end{subfigure}%
\hspace{\subplotgap}%
\begin{subfigure}[t]{\subplotwidth}
    \input{figures/swing_rate/vtab_structured}
\end{subfigure}%
\hspace{\subplotgap}%
\begin{subfigure}[t]{\subplotwidth}
    \input{figures/swing_rate/characters}
\end{subfigure}%
\hspace{\subplotgap}%
\begin{subfigure}[t]{\subplotwidth}
    \input{figures/swing_rate/fungi}
\end{subfigure}
\\[-3ex]

\resizebox{0.8\textwidth}{!}{\input{figures/swing_rate/legend}}

\caption{
\textbf{Stronger augmentation helps higher swing rates.}
Intermediate swing rates ($\frac{1}{4}$,$\frac{3}{4}$) mostly lie between MAE (swing rate of 0), \textbf{\textcolor{half_color}{Half-Swingers}} (swing rate of $\frac{1}{2}$), and \textbf{\textcolor{full_color}{Full-Swingers}} (swing rate of 1).
We choose these augmentation and color-masking settings because they are the 10-task frozen-feature winners at 200 pre-training epochs for MAE/ColorMAE (\plotmarker{green_weak}), Half-Swingers (\plotmarker{none_medium}) and Full-Swingers (\plotmarker{none_strong}).
}
\label{fig:swing_rate}
\end{figure*}

\textbf{Evaluation Setup: Fine-tuning.}
We fine-tune on IN-1K, the 21 datasets from OmniBenchmark \cite{zhang2022benchmarking}, chosen for its diversity of tasks, and ADE20K \cite{zhou2017scene} for segmentation.
For IN-1K, we fine-tune all 81 checkpoints, tuning the learning rate for each and report the best.
For all three algorithms, pre-training with green color masking and weak augmentations leads to the best IN-1K fine-tuning, so we initialize fine-tuning on OmniBenchmark and ADE20K using the checkpoints trained with these settings; we tune the learning rate on each dataset and checkpoint.
The OmniBenchmark datasets have different lengths so we fix the number of optimization steps across them.
For ADE20K, we use the EoMT segmentation method \cite{eomt}, which is performant and compute efficient.
Please see the supplementary Sec.~\ref{sec:appendix_finetuning} for details.

\textbf{Results: Swingers retains MAE's fine-tunability (Tab.~\ref{tab:main_ft}).}
In our controlled experiments, both Swinger variants fine-tune as well as MAE on IN-1K when pre-trained for 800 epochs.
When pre-trained for 100 epochs, Swingers provides a slight boost on IN-1K and a larger boost on OmniBenchmark: +1.7\% and +1.9\%, for Half-Swingers and Full-Swingers, respectively.
On ADE20K, Half-Swingers is comparable to the baseline at both pre-training lengths.


\subsection{Varying Swing Rate}
\label{sec:swing_rate}

\textbf{Pre-training Setup and Evaluations.}
Full-Swingers exchanges \cls{} tokens on every image and Half-Swingers on half of them, so we treat this ``swing rate'' as a knob.
We pre-train ViT-B/16 for 200 epochs on ImageNet-1K with swing rates \{0, $\frac{1}{4}$, $\frac{1}{2}$, $\frac{3}{4}$, 1\}, where 0 is our MAE baseline.
We cross swing rate with three settings: weak augmentation with green color masking, and medium and strong augmentation without color masking.
These are the winning settings at 200 pre-training epochs on the 10-task, frozen-feature evals for MAE, Half-Swingers and Full-Swingers, respectively.

\textbf{Results: Stronger augmentation calls for a higher swing rate (Fig.~\ref{fig:swing_rate}).}
On IN-1K, IN-21K, instance retrieval, iNat21-mini, and FungiTastic, the best swing rate is $\frac{1}{4}$ with weak augmentation and $\frac{3}{4}$ with strong augmentation.
With medium augmentation, Half-Swingers ($\frac{1}{2}$) performs best on every task except Omniglot.
This suggests that the less similar the views are, the more often Swingers should swing.
Omniglot is the exception, where accuracy tends to rise with swing rate in all three settings, e.g., from 46\% (MAE) to 87\% (Full-Swingers) with weak augmentation.

\subsection{Varying Mask Rate and Model Size}
\label{sec:mask_rate_and_model_size}

\textbf{Pre-training Setup and Evaluations.}
We pre-train Full-Swingers and Half-Swingers for 200 epochs on ImageNet-1K without color masking.
We run two sweeps, each crossing augmentation strength \{medium, strong\} with a second design axis to study their interaction: masking ratio \{55\%, 65\%, 75\%, 85\%\} using the ViT-Base size, and model size \{Tiny, Small, Base, Large\} using a 75\% masking ratio.
We use the same frozen-feature evaluations as in Sec.~\ref{sec:pre_train_aug_color_epochs}, and show the full results in supplementary Figs.~\ref{fig:mask_rate} and~\ref{fig:model_size}.

\textbf{Results: Lower mask rates help Full-Swingers (supplementary Fig.~\ref{fig:mask_rate}).}
Full-Swingers performs better with lower mask rates than MAE's default of 75\%.
At a 65\% mask rate, it reaches 50\% IN-1K $k$NN accuracy, matching Half-Swingers' best result.
Compared with the default 75\%, this lower mask rate also improves Full-Swingers on 10-shot IN-21K, all three mini-VTAB subsets, instance retrieval, animal re-identification, and fine-grained species classification.
The exception is Omniglot, where accuracy drops from 83\% to 73\%, which may be a worthwhile trade-off.
Stronger augmentation also makes Full-Swingers more robust to the choice of mask rate.
In contrast, Half-Swingers performs best or near-best at the default 75\% mask rate.
Omniglot is again the exception: raising the mask rate to 85\% improves accuracy +8\%/+12\% with medium/strong augmentation.

\textbf{Results: Full-Swingers wins at smaller ViT sizes and Half-Swingers at larger sizes (supplementary Fig.~\ref{fig:model_size}).}
Both variants generally improve with model size.
Full-Swingers outperforms Half-Swingers at smaller sizes, but Half-Swingers scales better to ViT-L.
This comparison uses the default 75\% mask rate, which is suboptimal for Full-Swingers, so a lower mask rate may narrow the gap at larger sizes.
Stronger augmentation tends to help Full-Swingers across sizes.
For Half-Swingers, the pattern is more mixed: stronger augmentation helps for ViT-Tiny and ViT-Large.

\subsection{State Probing for World Modeling}
\label{sec:wm}

\begin{table}[h!]
\centering
\providecommand{\lpmae}{MAE/ColorMAE}
\providecommand{\lphalf}{\textcolor{half_color}{\textbf{Half-Swingers}}}
\providecommand{\lpfull}{\textcolor{full_color}{\textbf{Full-Swingers}}}
\providecommand{\lphalfbg}{\rowcolor{half_color!10}}
\providecommand{\lpfullbg}{\rowcolor{full_color!10}}
\caption{
\textbf{Masked Swingers shows promise for world modeling.}
Masked Swingers improves state probing NMSE (lower $\downarrow$ is better) on the three MotionJEPA games~\cite{motionjepa}.
And this improvement is \emph{large}, e.g., 64\% error reduction from MAE/ColorMAE to our Full-Swingers (averaged over all tasks and models).
The MotionJEPA result is \emph{not} directly comparable:
its encoder is ViT-T/14 at $112^2$px trained from scratch on each game, whereas all other models are ViT-B/16 at $224^2$px pre-trained on IN-1K.
\textbf{Best} and \underline{2$^{nd}$ best}.
}
\label{tab:world_model}
\footnotesize
\setlength{\tabcolsep}{7pt}
\begin{tabular}{lccc}
Model & Pong & Dino & Golf \\
\toprule
\rowcolor{gray!12} \multicolumn{4}{l}{\textit{ViT-B/16 pre-trained by us on ImageNet-1K}} \\
MAE/ColorMAE (mean)   & 0.044 & 0.058 & 0.091 \\
MAE/ColorMAE (best)   & 0.007 & 0.031 & \underline{0.033} \\
\lphalfbg \textcolor{half_color}{\textbf{Half-Swingers (mean)}} & 0.012 & 0.032 & 0.054 \\
\lphalfbg \textcolor{half_color}{\textbf{Half-Swingers (best)}} & \underline{0.005} & \underline{0.013} & \textbf{0.024} \\
\lpfullbg \textbf{\textcolor{full_color}{Full-Swingers (mean)}} & 0.009 & 0.023 & 0.036 \\
\lpfullbg \textbf{\textcolor{full_color}{Full-Swingers (best)}} & \underline{0.005} & \textbf{0.012} & \textbf{0.024} \\
\midrule
\rowcolor{gray!12} \multicolumn{4}{l}{\textit{ViT-B/16 pre-trained by others on ImageNet-1K}} \\
SimDINOv2             & 0.006 & 0.044 & 0.059 \\
Bootleg               & 0.008 & 0.085 & 0.061 \\
LeJEPA                & 0.014 & 0.052 & 0.102 \\
ColorMAE (best of 3)  & 0.011 & 0.033 & 0.038 \\
MAE                   & 0.034 & 0.035 & 0.123 \\
\midrule
\rowcolor{gray!12} \multicolumn{4}{l}{\textit{ViT-T/14 trained by~\cite{motionjepa} on the games directly}} \\
MotionJEPA            & \textbf{0.004} & 0.021 & 0.061 \\
\bottomrule
\end{tabular}
\end{table}

\textbf{Masked Swingers reduces state-prediction error by 64\% relative to MAE.}
We probe frozen features for state information on three tasks, Pong, Dino, and Golf, using the public codebase of \citet{motionjepa}.
We evaluate our models from Sec.~\ref{sec:pre_train_aug_color_epochs} along with the reference models.
Our Swingers substantially outperforms the MAE/ColorMAE (Tab.~\ref{tab:world_model}), reducing error by 64\% averaged over all models and tasks.
It also beats SimDINOv2, reducing error by 73\% on Dino and 59\% on Golf.
These tasks require low-level features, such as the character's height or an obstacle's position.
Swingers' strength here is therefore consistent with its gains on other low-level tasks in Sec.~\ref{sec:pre_train_aug_color_epochs}, such as the mini-VTAB structured subset and character recognition.

\subsection{Analysis, Ablations, and Alternatives}
\label{sec:analysis_ablations_alternatives}

\begin{table}[h!]
  \centering
  \begin{minipage}[c]{0.51\linewidth}
  \captionsetup{justification=raggedright, singlelinecheck=false, skip=0pt}
  \caption{\textbf{Full-Swingers need stronger augmentation when pre-trained for longer.}
  Results are IN-1K $k$NN top-1 \% acc.
  ``Red'' masks larger contiguous areas, ``None'' is MAE-style.}
  \label{tab:augs_and_colors}
  \end{minipage}%
  \hfill
  \begin{minipage}[c]{0.48\linewidth}
  \centering
  \scriptsize
  \setlength{\tabcolsep}{2pt}
  \begin{tabular}{llccc}
    \multirow[b]{2}{*}[-1pt]{\makecell[l]{Aug.\\Strength}} & \multirow[b]{2}{*}[-1pt]{\makecell[l]{Color\\Masking}}
    & \multicolumn{3}{c}{Pre-train epochs} \\
    \cmidrule(lr){3-5}
    & & 100 & 200 & 800 \\
    \toprule
    Weak   & None & 39.1 & 33.9 & 33.2 \\
    Strong & Red  & 30.4 & 37.1 & 41.2 \\
    \bottomrule
  \end{tabular}
  \end{minipage}
\end{table}

\textbf{Full-Swingers can lose $k$NN accuracy on ImageNet with longer pre-training.}
We observe that Full-Swingers sometimes overfits to the pre-training task, so we show two illustrative examples.
Tab.~\ref{tab:augs_and_colors} shows how lengthening pre-training affects two settings at opposite ends of our design space:
\begin{enumerate*}[label=\textbf{(\roman*)}]
\item weak augmentation without color masking, which produces the most similar views, and
\item strong augmentation with red color masking, which produces the least similar views.
\end{enumerate*}
With longer pre-training, IN-1K $k$NN accuracy decreases when the views are too similar and increases when they are more dissimilar.
The mutual information between views thus shapes the representations our method learns.

\begin{table}[h!]
\centering
\caption{
\textbf{Swingers needs independently augmented views to learn ``higher-level'' frozen features.}
Encouraging view-\emph{specific} representations, by sharing augmentations between views, drops IN-1K $k$NN \textminus28\% and \textminus38\%, for Full-Swingers and Half-Swingers, respectively.
Yet it can boost downstream fine-tuning accuracy.
``\cls{}'' and ``Pool'' results are computed via $k$NN (IN-1K, Omniglot) and nearest-neighbor search (Retrieval, mAP).
\textbf{Best} and \underline{2$^{nd}$ best}.
}
\label{tab:same_aug}
\footnotesize
\setlength{\tabcolsep}{3.1pt}
\begin{tabular}{llrrrrrrrc}
\multirow[b]{2}{*}[-0.4pt]{\makecell[l]{Same\\Aug.}} &
\multirow[b]{2}{*}[-0.4pt]{\makecell[l]{Swing\\Strategy}} &
\multicolumn{3}{c}{ImageNet-1K} & \multicolumn{2}{c}{Retrieval} & \multicolumn{2}{c}{Omniglot} & \multirow[b]{2}{*}[-0.4pt]{\makecell[l]{Pre-train\\Loss}} \\
\cmidrule(r){3-5} \cmidrule(r){6-7} \cmidrule(){8-9}
& & \cls{} & Pool & FT & \cls{} & Pool & \cls{} & Pool & \\
\toprule
\multirow{3}{*}{Yes}
& None  & 38.0 & \underline{37.9} & 81.4 & 17.5 & 12.1 & 47.3 & \underline{40.8} & 0.397\\
& Half  & 12.1 & 29.6 & \underline{81.8} & 7.8 & 11.0 & 38.5 & 36.0 & \underline{0.359}\\
& Full  & 11.7 & 24.1 & \textbf{82.0} & 6.8 & 11.4 & 43.5 & 33.8 & \textbf{0.312}\\
\midrule
\multirow{3}{*}{No}
& None  & 35.3 & \textbf{38.1} & 81.3 & 15.4 & \underline{12.2} & 41.5 & 38.3 & 0.397\\
& Half  & \textbf{50.3} & 37.8 & 81.5 & \textbf{23.4} & 12.0 & \underline{68.0} & 37.5 & 0.396\\
& Full  & \underline{39.2} & 35.6 & 81.5 & \underline{17.7} & \textbf{12.9} & \textbf{83.3} & \textbf{47.3} & 0.392\\
\bottomrule
\end{tabular}
\end{table}
\begin{table*}[t!]
\centering
\caption{
\textbf{Swinging other tokens can also work well.}
Exchanging patch-pooled tokens may not need to be half-swung under this setting to perform well (i.e., 47\% ImageNet-1k $k$NN).
All 10 frozen-feature tasks, $k$NN on \cls{} and on pooled patches.
\textbf{Best} and \underline{2$^{nd}$ best} per column.
}
\label{tab:alternatives}
\scriptsize
\setlength{\tabcolsep}{2pt}
\resizebox{\linewidth}{!}{%
\begin{tabular}{l*{10}{rr}c}
& \multicolumn{2}{c}{IN-1K} & \multicolumn{2}{c}{IN-21K} & \multicolumn{2}{c}{Retrieval} & \multicolumn{2}{c}{ReID} & \multicolumn{2}{c}{iNat} & \multicolumn{2}{c}{Nat.} & \multicolumn{2}{c}{Spec.} & \multicolumn{2}{c}{Struct.} & \multicolumn{2}{c}{Omniglot} & \multicolumn{2}{c}{Fungi} & \multirow[b]{2}{*}[-1pt]{\makecell[l]{Pre-train\\Loss}} \\
\cmidrule(lr){2-3} \cmidrule(lr){4-5} \cmidrule(lr){6-7} \cmidrule(lr){8-9} \cmidrule(lr){10-11} \cmidrule(lr){12-13} \cmidrule(lr){14-15} \cmidrule(lr){16-17} \cmidrule(lr){18-19} \cmidrule(lr){20-21}
Setting & \cls{} & Pool & \cls{} & Pool & \cls{} & Pool & \cls{} & Pool & \cls{} & Pool & \cls{} & Pool & \cls{} & Pool & \cls{} & Pool & \cls{} & Pool & \cls{} & Pool &  \\
\toprule
exchange \cls{} (\textbf{\textcolor{full_color}{Full-Swingers}}) & 39.2 & 35.6 & 6.5 & 5.3 & 17.7 & 12.9 & 15.4 & 14.1 & 7.9 & 5.7 & 48.4 & 42.3 & 73.3 & 75.3 & 30.3 & 28.3 & \underline{83.3} & 47.3 & 15.0 & 15.6 & 0.392 \\
\hspace{1.2em}$\hookrightarrow$ on 50\% of imgs (\textbf{\textcolor{half_color}{Half-Swingers}}) & \underline{50.3} & 37.8 & \textbf{9.3} & 5.6 & \textbf{23.4} & 12.0 & \underline{17.8} & 13.2 & \underline{13.1} & 6.0 & \textbf{53.9} & 43.7 & \underline{76.4} & 75.3 & \textbf{31.3} & 28.0 & 68.0 & 37.5 & \textbf{22.9} & 16.3 & 0.396 \\
\midrule
exchange pooled patches & 42.8 & \textbf{46.8} & 6.7 & \textbf{8.4} & 19.6 & \textbf{15.6} & 16.7 & \textbf{17.0} & 9.1 & \textbf{10.3} & 47.7 & \textbf{53.8} & 76.3 & \textbf{77.3} & 30.3 & \textbf{31.7} & 60.0 & \textbf{80.5} & 19.4 & \textbf{20.4} & 0.394 \\
\hspace{1.2em}$\hookrightarrow$ on 50\% of imgs & 38.4 & \underline{44.8} & 5.7 & \underline{7.4} & 18.1 & 13.8 & 15.4 & 14.5 & 7.2 & \underline{8.8} & 43.8 & \underline{49.5} & 76.2 & \underline{76.9} & 30.1 & \underline{30.2} & 51.5 & 55.0 & 17.8 & \underline{20.1} & 0.396 \\
exchange \cls{} and pooled patches & 36.3 & 36.8 & 5.9 & 5.7 & 15.9 & \underline{14.8} & 14.5 & \underline{15.0} & 7.7 & 5.5 & 45.4 & 45.2 & 69.8 & 74.0 & 29.8 & 28.0 & 80.0 & \underline{64.0} & 13.7 & 13.6 & 0.392 \\
\hspace{1.2em}$\hookrightarrow$ on 50\% of imgs & \textbf{50.5} & 39.1 & 9.1 & 5.9 & 22.3 & 12.5 & 16.7 & 13.4 & \textbf{13.2} & 6.4 & \underline{53.8} & 44.9 & 76.1 & 75.2 & \underline{31.2} & 28.6 & 59.5 & 47.5 & \textbf{22.9} & 16.4 & 0.396 \\
give both \cls{} to both decoder views & 43.0 & 36.1 & 7.5 & 5.4 & 19.3 & 12.3 & 17.0 & 13.5 & 9.5 & 5.6 & 50.9 & 43.1 & 74.6 & 75.4 & 29.9 & 28.0 & \textbf{86.8} & 42.5 & 17.1 & 15.9 & 0.392 \\
\hspace{1.2em}$\hookrightarrow$ on 50\% of imgs & 48.5 & 38.1 & \underline{9.2} & 5.7 & \underline{22.4} & 12.2 & \textbf{19.2} & 13.4 & 12.3 & 6.0 & 53.6 & 43.8 & \textbf{76.6} & 76.2 & 31.1 & 28.1 & 78.3 & 42.5 & \underline{22.2} & 16.4 & 0.396 \\
give all encoder outputs to both decoder views & 20.9 & 27.9 & 2.2 & 3.6 & 11.7 & 11.1 & 12.8 & 12.8 & 2.6 & 4.0 & 28.3 & 36.4 & 72.3 & 75.0 & 26.4 & 27.5 & 30.3 & 32.5 & 9.9 & 13.8 & \textbf{0.358} \\
\hspace{1.2em}$\hookrightarrow$ on 50\% of imgs & 30.1 & 32.1 & 3.9 & 4.4 & 14.5 & 12.2 & 14.3 & 13.3 & 5.1 & 5.0 & 36.8 & 38.8 & 73.3 & 75.4 & 27.7 & 27.4 & 33.8 & 38.5 & 15.1 & 15.6 & \underline{0.382} \\
\bottomrule
\end{tabular}
}
\end{table*}
\providecommand{\lpchg}[2]{\textcolor{#1}{#2\%}}
\colorlet{lpup}{green!55!black}
\colorlet{lpdown}{red!80!black}
\begin{table*}[t!]
\centering
\scriptsize
\caption{
\textbf{Exchanging encoded \cls{} tokens between similar but distinct samples boosts accuracy for little cost.}
Starting from a Half-Swingers model pre-trained for 800 epochs, we continue pre-training for 50 epochs---increasing total compute cost by \texttimes1.06.
Instead of swinging across the same sample augmented differently, we swing across different, yet similar images (nearest neighbours over IN-1K train).
}
\label{tab:nn}
{\scriptsize
\setlength{\tabcolsep}{3pt}
\begin{tabular}{l c@{\hspace{2.5pt}}l c@{\hspace{2.5pt}}l c@{\hspace{2.5pt}}l c@{\hspace{2.5pt}}l c@{\hspace{2.5pt}}l c@{\hspace{2.5pt}}l c@{\hspace{2.5pt}}l c@{\hspace{2.5pt}}l c@{\hspace{2.5pt}}l c@{\hspace{2.5pt}}l}
 & \multicolumn{2}{c}{\makecell{IN-1K\\$k$NN}}
 & \multicolumn{2}{c}{\makecell{IN-21K\\10-shot $k$NN}}
 & \multicolumn{2}{c}{\makecell{iNat21-mini\\$k$NN}}
 & \multicolumn{2}{c}{\makecell{FungiTastic\\$k$NN}}
 & \multicolumn{2}{c}{\makecell{mini-VTAB\\Natural}}
 & \multicolumn{2}{c}{\makecell{mini-VTAB\\Specialized}}
 & \multicolumn{2}{c}{\makecell{mini-VTAB\\Structured}}
 & \multicolumn{2}{c}{\makecell{Instance\\Retrieval}}
 & \multicolumn{2}{c}{\makecell{Wildlife\\Re-ID}}
 & \multicolumn{2}{c}{\makecell{Omniglot\\20-way 1-shot}} \\
\toprule
Before & 56.07 & \multirow{2}{*}{\lpchg{lpup}{+5.27}} & 11.03 & \multirow{2}{*}{\lpchg{lpup}{+1.99}} & 15.11 & \multirow{2}{*}{\lpchg{lpup}{+6.04}} & 23.56 & \multirow{2}{*}{\lpchg{lpup}{+4.28}} & 57.40 & \multirow{2}{*}{\lpchg{lpup}{+3.04}} & 77.48 & \multirow{2}{*}{\lpchg{lpup}{+0.55}} & 34.39 & \multirow{2}{*}{\lpchg{lpup}{+1.38}} & 25.61 & \multirow{2}{*}{\lpchg{lpup}{+1.40}} & \textbf{18.84} & \multirow{2}{*}{\lpchg{lpdown}{\textminus0.45}} & \textbf{77.25} & \multirow{2}{*}{\lpchg{lpdown}{\textminus7.25}} \\
After & \textbf{61.34} &  & \textbf{13.02} &  & \textbf{21.15} &  & \textbf{27.84} &  & \textbf{60.44} &  & \textbf{78.03} &  & \textbf{35.77} &  & \textbf{27.01} &  & 18.39 &  & 70.00 &  \\
\bottomrule
\end{tabular}}
\end{table*}

\textbf{Swingers needs independent augmentation for strong frozen features (Tab.~\ref{tab:same_aug}).}
We ablate independent augmentation by giving both views the same augmentation, so they differ only in their masks.
This substantially hurts both variants, reducing IN-1K $k$NN accuracy 28\% for Full-Swingers and 38\% for Half-Swingers.
Shared augmentation also hurts Omniglot and retrieval.
These results show that independent augmentation encourages more view-agnostic or ``semantic'' representations.
Surprisingly, Full-Swingers \emph{without} independent augmentation achieves our best IN-1K fine-tuning accuracy (82.0\%) under these 200-epoch pre-training settings.
Using shared augmentation, the exchanged \cls{} token describes a view that differs only in its mask, which encourages a view-\emph{specific} summary that suits fine-tuning.

\textbf{We can also exchange pooled patch tokens (Tab.~\ref{tab:alternatives}).}
We test exchanging tokens other than the \cls{} token.
In one alternative, each decoder receives its own view's \cls{} token and patch tokens and the mean-pooled patch token from the other view.
This is competitive with exchanging \cls{} tokens, showing that the benefit comes from passing a cross-view summary rather than from the \cls{} token specifically.
We also exchange both the pooled patch token and the \cls{} token.
This combination works best when exchanging occurs on only 50\% of samples, as in Half-Swingers.

\textbf{An information bottleneck is crucial.}
The second last row in Tab.~\ref{tab:alternatives} provides \emph{all} encoded tokens to both decoder views, making it similar to CropMAE \cite{eymael2024efficient} (see Sec. \ref{sec:related_work} for discussion).
This ablation achieves 9\% lower reconstruction loss, due to more information transfer between views, yet it is 18\% worse than Full-Swingers on IN-1K $k$NN.
Providing all encoded tokens on half of the images (last row) improves 9\%, however, it is still 20\% worse than Half-Swingers.

\subsection{Post-training via Swinging Nearest Neighbors}
\label{sec:nn}

\textbf{Nearest-neighbor Swinging boosts IN-1K $k$NN +5.3\% with only 1.06$\times$ additional computational cost.}
We now experiment with Swinging the \cls{} tokens between \emph{different} images.
We take our best Swinger model from Sec. \ref{sec:pre_train_aug_color_epochs}, according to IN-1K $k$NN, and continue pre-training it by exchanging encoded \cls{} tokens between an image and the image of its nearest neighbor in the IN-1K training set (measured by cosine similarity and re-computed after each epoch).
We train for an additional 50 epochs (6\% extra from the initial 800 epochs of pre-training), using a 2e\textminus5 learning rate, 256 batch size, and minimal augmentation (random crop and flip).
This model significantly improves on most tasks, e.g.,
absolute gains of +5.3\% on IN-1K, +2.0\% on 10-shot IN-21K, and +6.0\% on iNat21-mini (Tab. \ref{tab:nn}).
However, this model is slightly worse on re-identifying individual animals and significantly worse on Omniglot character recognition.
Swinging neighbors thus encourages the extraction of ``higher-level'' features over instance-specific features.

\subsection{Measuring View Invariance}
\label{sec:infonce}

\begin{figure}[h!]
\centering
\input{figures/infonce/figure}
\caption{
\textbf{Swingers can lower InfoNCE loss \cite{infonce} without optimizing for it.}
We compute InfoNCE over the ImageNet-1K validation set; positives are different augmentations of the same instance and negatives are the other 49,999 instances.
Each SSL algorithm has 27 points, varying in pre-training length (darker: pre-trained for longer), data augmentation, and color masking.
}
\label{fig:infonce}
\end{figure}

\textbf{Swingers encourages view invariance as measured by InfoNCE \cite{infonce}.}
To check if Swingers achieves view invariance, we compute the InfoNCE loss over the ImageNet-1K validation set where positives are different augmentations of matching instances and negatives are the other 49,999 non-matching instances.
Lower loss means a model's \cls{} is closer to its positives and/or farther from its negatives.
We compute this loss for all 81 pre-trained checkpoints from Sec. \ref{sec:pre_train_aug_color_epochs}.
Full-Swingers achieves 7.8/5.8, Half-Swingers achieves 7.1/5.2, and MAE/ColorMAE achieves 10.4/9.0 InfoNCE loss for the average/best checkpoints.
InfoNCE loss is the most common contrastive-learning objective, chosen to promote view invariance \cite{khosla2020supervised, radford2021learning, chen2020simple, he2020momentum}.
To optimize for it, large batches must be held in memory simultaneously---Swingers can learn it indirectly without requiring in-batch negatives or batch-wise statistics.
Fig. \ref{fig:infonce} plots all 81 checkpoints with InfoNCE loss (x-axis) and the 10-task frozen-feature average (y-axis).
The two axes are highly correlated, showing the invariance measure maps to downstream-task performance.
The highest correlations between downstream-task accuracy and InfoNCE loss are for instance retrieval and animal re-identification; the lowest correlations are for mini-VTAB Specialized, mini-VTAB Structured, and IN-1K.

\section{Related Work: Masked Modeling}
\label{sec:related_work}

\textbf{Cross-View Autoencoding.}
The closest method to ours is CropMAE~\cite{eymael2024efficient}, designed for video feature-correspondence tasks such as DAVIS~\cite{ponttuset20182017davischallengevideo}.
CropMAE makes a second view with a mild crop and a random horizontal flip, then masks nearly all of it ($\rho$ = 98.5\%).
It encodes the unmasked original view and the masked second view separately, then decodes the masked patches using all tokens from both views.
CropMAE paper does not report IN-1K, so we evaluate its pre-trained ViT-S checkpoint ourselves: it reaches 12\% IN-1K $k$NN accuracy, compared with 40\% for our Swingers ViT-S (supplementary Fig.~\ref{fig:model_size}).
CroCo~\cite{weinzaepfel2022croco} and SiamMAE~\cite{gupta2023siamese} are the next closest methods, but they obtain their views from data rather than augmentation: CroCo uses different views of a 3D scene and SiamMAE uses neighboring video frames.
CroCo and CroCo v2~\cite{weinzaepfel2023croco} reach 19\% and 29\% ViT-B IN-1K $k$NN accuracy, though they pre-train on different data for different lengths, so these numbers are not directly comparable to ours.
Unlike these methods, which let the decoder access every token of the other view, we pass \textbf{\emph{one}} token between views.
This information bottleneck, combined with strong, independent augmentation and masking of each view, encourages a view-agnostic input summary.

\textbf{Other Masked Autoencoders.}
Many papers, including ColorMAE~\cite{hinojosa2024colormae}, invent clever masking strategies~\cite{li2022semmae, kakogeorgiou2022hide, shi2022adversarial, shin2024self, choi2024salience, wang2023hard, feng2023evolved, yang2026pursuit, li2022uniform, liu2023good, haghighat2024pretraining}.
Others create new image-derived targets~\cite{wei2022masked, xie2023masked, wang2023masked, liu2023pixmim}.
And some papers improve the decoder~\cite{fu2025rethinking, rajasegaran2025gaussianmaskedautoencoders, wang2023masked}.
These may provide complementary improvements to our Swingers. 
Since masking affects the mutual information between views and could thus swing with us, we thoroughly experiment with ColorMAE's masking (Sec.~\ref{sec:pre_train_aug_color_epochs}).
We choose ColorMAE over alternatives because it is data-independent, model-free, and intuitive.

\textbf{Masked Latent Modeling.}
There is a related but distinct SSL framework which masks inputs, then predicts representations (not pixels, thus are not autoencoders) at the masked locations~\cite{assran2023self, lowe2026self, wei2024towards, baevski2022data2vec, zhou2021ibot}.
The latents are computed from a ``teacher'' model that is the EMA of the ``student'', thus relies on self-distillation.
Our doors are open to incorporating these latent-prediction methods into our Swingers framework.

\section{Discussion: Swinging Shut, For Now...}

\textbf{Limitations and future work.}
The main limitation of Masked Swingers is that it requires data augmentation to make different views, which is not required by MAE.
However, data augmentation is ubiquitous across deep-learning applications, so this limitation is minor in practice.
Swingers can underperform SOTA methods from other SSL frameworks, e.g., SimDINOv2.
However, we narrow this gap while keeping the strengths of the autoencoding framework, i.e., robustness to training settings (by predicting stationary/grounded targets, e.g., images) and low device-memory requirements (by avoiding cross-sample interactions).
And we achieve this with a simple and intuitive idea.
The main limitation of our study is that we only pre-train on natural images.
We choose this for focus, leaving other experiments to future swinging in space (3D data) and time (video).

\textbf{Conclusion.}
We introduce a novel autoencoding algorithm that encourages view-agnostic representations by exchanging encoded \cls{} tokens between two different views of an image before decoding.
Via thorough controlled experiments, we show our Masked Swingers significantly outperforms the MAE on fine-grained tasks.
Through ablations we show that independent augmentations and an information bottleneck are both crucial to extracting ``higher-level'' features.
We further show our Swingers' potential as a frame-encoder in world modeling.
Join us in swinging tokens to encourage view-agnostic representations via autoencoding.

\section{LLM Usage Statement}
We used Claude (Anthropic), via Claude Code.
We, the authors, wrote almost all words of the main text; Claude suggested a few clarifications.
Claude wrote all the words of the appendix, except for Sec. \ref{sec:appendix_effrank} and the figure captions.
To do so faithfully, we gave it access to our full codebase and results files; we double checked it.
Claude double checked all results in the main paper and appendix for consistency with our results files.
Claude did most of the coding; we take responsibility for it.
No AI had any input on the method, research questions, experimental design, or overall exposition.

\iftoggle{cvprfinal}{
\section*{Acknowledgements}
AF did much of this work during his PhD, which was funded by an NSERC PGS-D.
All pre-training runs were performed on Google TPUs through the TRC program.
ES and GWT are supported by Canada CIFAR AI Chairs.
Resources used in preparing this research were provided, in part, by the Province of Ontario, the Government of Canada through CIFAR, and \href{https://vectorinstitute.ai/partnerships/current-partners/}{companies sponsoring} the Vector Institute.
}{}

{
    \small
    \bibliographystyle{ieeenat_fullname}
    \bibliography{references}
}

\clearpage
\setcounter{page}{1}
\maketitlesupplementary
\appendix

\urldef{\codeurl}\url{https://github.com/antofuller/masked_swingers}
\iftoggle{cvprfinal}{
We release pre-training code and pre-trained models:
\codeurl.
}{
We will release pre-training code and pre-trained models upon publication.
}

\section{Additional pre-training details}
\label{sec:appendix_pretraining}

We implement pre-training in JAX and train on TPUs.
Each step samples 1,024 ImageNet-1K training images and makes two views of each (2,048 views per step), so our MAE baseline runs the same two-view pipeline, with each decoder receiving its own view's \cls{} token.
We augment each view independently with a random resized crop (scale $[0.08, 1]$, bicubic), a random horizontal flip, and RandAugment~\cite{cubuk2020randaugment}, where weak/medium/strong apply 1/2/4 operations at magnitude 5/9/15 (magnitude noise std 0.5); we use no color jitter or random erasing.
We then mask each view independently ($\rho{=}75\%$ unless stated otherwise).
For color masking, we generate noise on the $14 \times 14$ patch grid for each view, filtering white Gaussian noise with a Gaussian ($\sigma{=}2$ patches) for red or a difference of Gaussians ($\sigma{=}1$ and $4$ patches) for green, and we keep the lowest-scoring patches.
The encoder and decoder use fixed 2D sin-cos position embeddings and no dropout or stochastic depth.
The decoder has 8 blocks with head dimension 32 and width 128/256/512/512 for ViT-T/S/B/L encoders, where ViT-L has 24 blocks of width 1024 with 16 heads.
We use AdamW ($\beta_1{=}0.9$, $\beta_2{=}0.95$, weight decay 0.05 on weight matrices only) with a peak learning rate of $6{\times}10^{-4}$ ($1.5{\times}10^{-4} \times 1024/256$).
The learning rate warms up linearly from $10^{-6}$ over 10/20/40 epochs for 100/200/800-epoch runs, then follows a cosine decay to $10^{-5}$; we use no gradient clipping.

\section{Additional fine-tuning details}
\label{sec:appendix_finetuning}

\subsection{ImageNet-1K fine-tuning details}

We fine-tune on IN-1K following the ViT-B recipe of \citet{he2022masked}.
We train for 100 epochs with a batch size of 1,024, using AdamW ($\beta_1{=}0.9$, $\beta_2{=}0.999$, weight decay 0.05 on weight matrices only), layer-wise learning-rate decay of 0.65, and drop path of 0.1.
The learning rate warms up linearly from 0 over 5 epochs, then follows a cosine decay to $10^{-6}$.
For augmentation, we use a random resized crop (scale $[0.08, 1]$), a horizontal flip, RandAugment~\cite{cubuk2020randaugment} (2 operations, magnitude 9), random erasing ($p{=}0.25$), and either Mixup ($\alpha{=}0.8$) or CutMix ($\alpha{=}1.0$), chosen per batch with equal probability, with label smoothing of 0.1.
For each pre-trained model, we sweep the base learning rate over $\{2.5, 5, 10, 20\} {\times} 10^{-4}$ (peak learning rate $=$ base $\times\, 1024/256$) and report the best.
We evaluate on the 50K IN-1K validation images, resized to 256 pixels and center-cropped to $224 \times 224$.

\subsection{OmniBenchmark fine-tuning details}
\label{sec:appendix_omnibench}

We fine-tune on OmniBenchmark V2~\cite{zhang2022benchmarking}, which spans 21 realms (\eg, birds, aircraft, food, and activities), each with 20 to 750 classes and 1.9K to 97K training images.
We fine-tune separately on each realm, training the full encoder together with a linear head on the final-layer \cls{} token.
We train for 1,000 steps with a batch size of 256 (2.6 to 143 epochs, depending on the realm's size), using AdamW ($\beta_1{=}0.9$, $\beta_2{=}0.999$, weight decay 0.05 on weight matrices only) and bf16 mixed precision; we use no layer-wise learning-rate decay, drop path, Mixup, CutMix, or label smoothing.
The learning rate warms up linearly from 0 over the first 50 steps, then follows a cosine decay to 1\% of its peak.
For augmentation, we use a random resized crop (scale $[0.08, 1]$, bicubic), a horizontal flip, and RandAugment~\cite{cubuk2020randaugment} (2 operations, magnitude 9, magnitude noise std 0.5).
For each pre-trained model and realm, we sweep the learning rate over $\{0.3, 1, 3, 10, 30\} {\times} 10^{-4}$, select it by accuracy on the realm's validation split, and report that run's test accuracy, averaged over the 21 realms.
We evaluate with images resized to 256 pixels and center-cropped to $224 \times 224$.

\subsection{ADE20K fine-tuning details}
\label{sec:appendix_ade20k}

We fine-tune on ADE20K~\cite{zhou2017scene} semantic segmentation with EoMT~\cite{eomt}, using the official code and its ADE20K recipe: $512 \times 512$ crops, a batch size of 16, 31 epochs (${\sim}$39K iterations), AdamW with layer-wise learning-rate decay of 0.8, and Mask2Former-style~\cite{cheng2022masked} losses.
Following EoMT's ViT-B configurations, the 100 learnable queries join the last 3 of the encoder's 12 blocks, and we anneal masked attention away block by block between steps 6,520 and 32,600.
We initialize the encoder from the pre-trained weights and bicubically interpolate its sin-cos position embeddings from $14 \times 14$ to $32 \times 32$.
For each pre-trained model, we sweep the learning rate over $\{0.5, 1, 2, 4\} {\times} 10^{-4}$ for 800-epoch models and $\{1, 2, 4, 8\} {\times} 10^{-4}$ for 100-epoch models, which prefer higher learning rates, and report the best.
We report single-scale, sliding-window mIoU on the 2,000 validation images after the final epoch.

\section{Reference models}
\label{sec:appendix_references}

All reference models are ViT-B/16 pre-trained on IN-1K at $224 \times 224$ pixels.
We download their public weights and evaluate them with the same frozen-feature protocols as our models.
For MAE~\cite{he2022masked}, we use the official 1600-epoch checkpoint, loaded through timm as \texttt{vit\_base\_patch16\_224.mae}.
For ColorMAE~\cite{hinojosa2024colormae}, we use the official green-noise checkpoints pre-trained for 300, 800, and 1600 epochs (\url{https://huggingface.co/carlosh93/colormae}).
LeJEPA~\cite{balestriero2025lejepa} has no official ViT-B/16 release, so we use the 100-epoch checkpoint of a public reproduction (\url{https://huggingface.co/OK-AI/lejepa-vitb16-pretrain-in1k}).
For SimDINOv2~\cite{wu2025simplifying}, we use the official 100-epoch checkpoint (\url{https://github.com/RobinWu218/SimDINO}).
For Bootleg~\cite{lowe2026self}, we use the 600-epoch checkpoint shared with us by the authors directly.
We evaluate its EMA target encoder and discard its 4 register tokens before pooling.
For the comparisons in Sec.~\ref{sec:related_work}, we use two further sets of official checkpoints.
The first is the IN-1K-pre-trained ViT-S/16 CropMAE checkpoint (\url{https://github.com/alexandre-eymael/CropMAE}).
The second is the ViT-B CroCo and CroCo v2 checkpoints (\url{https://github.com/naver/croco}); these pre-train on image pairs of 3D scenes rather than IN-1K.

\section{Frozen-feature evaluation details}
\label{sec:appendix_frozen}

We resize each image's shorter side to 224 pixels, center-crop to $224 \times 224$, and extract the final-layer \cls{} token and the mean of the final-layer patch tokens.
For $k$NN, we standardize each feature dimension using training-set statistics, L2-normalize, and weight each of the $k$ nearest training neighbors' votes by $\exp(s/T)$, where $s$ is the cosine similarity.
We sweep $k$ and $T$ jointly, starting from a log-spaced $12 \times 12$ grid ($k {\in} [1, 200]$, $T {\in} [0.005, 0.5]$).
Whenever the best value falls on an edge of the grid, we extend that axis until the best $(k, T)$ is interior.
We cap $k$ at 1,000.
We select $(k, T)$ on the evaluation split, identically for every model.
For instance retrieval, animal re-identification, and Omniglot, we rank by cosine similarity between L2-normalized features, with no tuning.
For ROxford5k, RParis6k, and ILIAS, we crop each query to its annotated bounding box; for Omniglot, we assign each test image the class of its nearest support image.

\section{Varying Mask Rate and Model Size, alongside Augmentation}
\label{sec:appendix_mask_rate_model_size}

Figs.~\ref{fig:mask_rate} and~\ref{fig:model_size} show the full results of the mask rate and model size sweeps, which we discuss in Sec.~\ref{sec:mask_rate_and_model_size} of the main paper.

\begin{figure*}[t!]
\centering

\begin{subfigure}[t]{\subplotwidth}
    \input{figures/mask_rate/imagenet_knn}
\end{subfigure}%
\hspace{\subplotgap}%
\begin{subfigure}[t]{\subplotwidth}
    \input{figures/mask_rate/imagenet_21k_knn}
\end{subfigure}%
\hspace{\subplotgap}%
\begin{subfigure}[t]{\subplotwidth}
    \input{figures/mask_rate/instance_retrieval}
\end{subfigure}%
\hspace{\subplotgap}%
\begin{subfigure}[t]{\subplotwidth}
    \input{figures/mask_rate/animal_reidentification}
\end{subfigure}%
\hspace{\subplotgap}%
\begin{subfigure}[t]{\subplotwidth}
    \input{figures/mask_rate/inat}
\end{subfigure}
\\[-0.5ex]
\begin{subfigure}[t]{\subplotwidth}
    \input{figures/mask_rate/vtab_natural}
\end{subfigure}%
\hspace{\subplotgap}%
\begin{subfigure}[t]{\subplotwidth}
    \input{figures/mask_rate/vtab_specialized}
\end{subfigure}%
\hspace{\subplotgap}%
\begin{subfigure}[t]{\subplotwidth}
    \input{figures/mask_rate/vtab_structured}
\end{subfigure}%
\hspace{\subplotgap}%
\begin{subfigure}[t]{\subplotwidth}
    \input{figures/mask_rate/characters}
\end{subfigure}%
\hspace{\subplotgap}%
\begin{subfigure}[t]{\subplotwidth}
    \input{figures/mask_rate/fungi}
\end{subfigure}
\\[-3ex]

\resizebox{0.8\textwidth}{!}{\input{figures/mask_rate/legend}}

\caption{
\textbf{\textcolor{full_color}{Full-Swingers} prefers lower mask rates than \textcolor{half_color}{Half-Swingers}.}
For example, at a 65\% mask rate, our Full-Swingers ties or outperforms Half-Swingers at its best mask rate of 75\%.
Omniglot is an exception where higher mask rates improves Full-Swingers.
Stronger augmentation tends to make Full-Swingers more robust to mask rates and it tends to hurt Half-Swingers.
}
\label{fig:mask_rate}
\end{figure*}
\begin{figure*}[t!]
\centering

\begin{subfigure}[t]{\subplotwidth}
    \input{figures/model_size/imagenet_knn}
\end{subfigure}%
\hspace{\subplotgap}%
\begin{subfigure}[t]{\subplotwidth}
    \input{figures/model_size/imagenet_21k_knn}
\end{subfigure}%
\hspace{\subplotgap}%
\begin{subfigure}[t]{\subplotwidth}
    \input{figures/model_size/instance_retrieval}
\end{subfigure}%
\hspace{\subplotgap}%
\begin{subfigure}[t]{\subplotwidth}
    \input{figures/model_size/animal_reidentification}
\end{subfigure}%
\hspace{\subplotgap}%
\begin{subfigure}[t]{\subplotwidth}
    \input{figures/model_size/inat}
\end{subfigure}
\\[-0.5ex]
\begin{subfigure}[t]{\subplotwidth}
    \input{figures/model_size/vtab_natural}
\end{subfigure}%
\hspace{\subplotgap}%
\begin{subfigure}[t]{\subplotwidth}
    \input{figures/model_size/vtab_specialized}
\end{subfigure}%
\hspace{\subplotgap}%
\begin{subfigure}[t]{\subplotwidth}
    \input{figures/model_size/vtab_structured}
\end{subfigure}%
\hspace{\subplotgap}%
\begin{subfigure}[t]{\subplotwidth}
    \input{figures/model_size/characters}
\end{subfigure}%
\hspace{\subplotgap}%
\begin{subfigure}[t]{\subplotwidth}
    \input{figures/model_size/fungi}
\end{subfigure}
\\[-3ex]

\resizebox{0.8\textwidth}{!}{\input{figures/mask_rate/legend}}

\caption{
\textbf{\textcolor{full_color}{Full-Swingers} and \textcolor{half_color}{Half-Swingers} tend to improve when increasing model size.}
The gain is monotonic for Omniglot but is more complex for other tasks.
For example, Full-Swingers (aug=medium) peaks on animal re-identification at ViT-Small, then decreases for ViT-Base and ViT-Large; increasing augmentation helps for these larger architectures.
}
\label{fig:model_size}
\end{figure*}

\section{Effective Rank of the \cls{} Token}
\label{sec:appendix_effrank}

\begin{figure*}[t]
  \centering
  \begin{minipage}[c]{0.32\linewidth}
    \caption{
    \textbf{Swingers distributes the variation of the \cls{} token more evenly across directions.}
    Effective rank of the 768-d \cls{} covariance on IN-1K val vs.\ (left) 10-shot IN-21K $k$NN accuracy and (right) Omniglot 20-way one-shot acc., for 81 pre-trained models (3 aug strengths \texttimes{} 3 mask types \texttimes{} 3 training lengths \texttimes{} 3 algorithms).
    }
    \label{fig:effrank}
  \end{minipage}\hfill
  \begin{minipage}[c]{0.66\linewidth}
    \centering
    \resizebox{\linewidth}{!}{\input{figures/rank_acc/legend}}
    \begin{minipage}[t]{0.485\linewidth}
      \centering
      \input{figures/rank_acc/effrank_cov_in21k}
    \end{minipage}\hfill
    \begin{minipage}[t]{0.485\linewidth}
      \centering
      \input{figures/rank_acc/effrank_cov_omniglot}
    \end{minipage}
  \end{minipage}
\end{figure*}

\textbf{Swingers' \cls{} token varies along more directions.}
In MAE, the decoder predicts masked patches from the visible patch tokens of the same view, so the \cls{} token is not directly encouraged to carry information.
In Swingers, the \cls{} token is the only information about the other view that the decoder receives, so the more descriptive it is, the lower the reconstruction loss.
To measure this, we compute the effective rank~\cite{roy2007effective} of the \cls{} covariance on the IN-1K validation set, which measures how evenly the variation across images is spread over the embedding's directions: it equals the embedding dimension when every direction contributes equally, and approaches 1 when a single direction dominates.
Fig.~\ref{fig:effrank} plots effective rank against 10-shot IN-21K and Omniglot accuracy for all 81 models from Sec.~\ref{sec:pre_train_aug_color_epochs} of the main paper.
On IN-21K, accuracy rises with rank up to a point and then declines, whereas on Omniglot it rises steadily with rank.
One possible explanation is that the additional directions capture ``lower-level'' features, such as the subtle stroke shapes that distinguish Omniglot's 1,623 handwritten characters, which help Omniglot but eventually hurt IN-21K.

\section{All Results}
\label{sec:appendix_all_results}

For completion, we present results tables that include all 81 pre-trained models from Sec.~\ref{sec:pre_train_aug_color_epochs} of the main paper.
We also present results of evaluations that we run using official/public checkpoints.

\clearpage
\onecolumn

\vspace*{\fill}
\begingroup
\scriptsize
\providecommand{\baselineColSep}{2pt}  
\setlength{\tabcolsep}{\baselineColSep}
\par\noindent\makebox[\linewidth][c]{%
\begin{tabular}{lrrrrrrrrrrrrrrrrrrrr}
\multicolumn{1}{l}{} & \multicolumn{1}{l}{} & \multicolumn{1}{l}{} & \multicolumn{2}{c}{IN-1K} & \multicolumn{2}{c}{\makecell{IN-21K\\10-shot}} & \multicolumn{2}{c}{\makecell{m-VTAB\\Natural}} & \multicolumn{2}{c}{\makecell{m-VTAB\\Specialized}} & \multicolumn{2}{c}{\makecell{m-VTAB\\Structured}} & \multicolumn{8}{c}{\makecell{Instance Retrieval\\mAP}} \\
\cmidrule(lr){4-5} \cmidrule(lr){6-7} \cmidrule(lr){8-9} \cmidrule(lr){10-11} \cmidrule(lr){12-13} \cmidrule(lr){14-21}
\multicolumn{1}{l}{} & \multicolumn{1}{l}{} & \multicolumn{1}{l}{\multirow[b]{2}{*}[-\cmidruleheight]{\makecell{Pre-train\\GFLOPs}}} & \multicolumn{2}{c}{kNN} & \multicolumn{2}{c}{kNN} & \multicolumn{2}{c}{kNN} & \multicolumn{2}{c}{kNN} & \multicolumn{2}{c}{kNN} & \multicolumn{2}{c}{ROxford} & \multicolumn{2}{c}{RParis} & \multicolumn{2}{c}{ILIAS} & \multicolumn{2}{c}{SOP} \\
\cmidrule(lr){4-5} \cmidrule(lr){6-7} \cmidrule(lr){8-9} \cmidrule(lr){10-11} \cmidrule(lr){12-13} \cmidrule(lr){14-15} \cmidrule(lr){16-17} \cmidrule(lr){18-19} \cmidrule(lr){20-21}
Model & Ep. & \multicolumn{1}{l}{} & CLS & Pool & CLS & Pool & CLS & Pool & CLS & Pool & CLS & Pool & CLS & Pool & CLS & Pool & CLS & Pool & CLS & Pool \\
\midrule
Bootleg & 600 & 4.15 & 72.6 & 64.8 & 15.9 & 12.3 & 61.9 & 56.6 & 77.2 & 78.0 & 27.2 & 27.9 & 13.3 & 9.3 & 40.5 & 27.7 & 10.7 & 5.0 & 34.9 & 33.2 \\
LeJEPA & 100 & 2.01 & 61.1 & 51.9 & 12.1 & 9.1 & 57.6 & 52.7 & 79.5 & 78.8 & 28.4 & 30.1 & 12.9 & 13.8 & 34.1 & 32.6 & 7.8 & 5.7 & 33.7 & 32.6 \\
ColorMAE & 300 & 1.08 & 42.7 & 45.7 & 6.3 & 7.3 & 45.2 & 49.9 & 74.9 & 74.0 & 29.5 & 28.7 & 8.9 & 6.3 & 23.1 & 15.1 & 2.9 & 2.1 & 30.1 & 20.9 \\
ColorMAE & 800 & 2.87 & 47.3 & 51.9 & 6.7 & 8.5 & 45.2 & 52.6 & 74.9 & 73.8 & 29.4 & 29.1 & 5.9 & 5.4 & 20.9 & 13.9 & 2.6 & 1.9 & 29.1 & 19.2 \\
ColorMAE & 1600 & 5.74 & 53.8 & 55.1 & 8.0 & 9.2 & 47.8 & 53.8 & 75.4 & 74.3 & 31.0 & 29.7 & 7.2 & 5.3 & 20.9 & 13.1 & 3.4 & 1.8 & 29.6 & 19.2 \\
\bottomrule
\end{tabular}
}%
\par
\endgroup

\vspace*{\fill}
\begingroup
\scriptsize
\providecommand{\baselineColSep}{2pt}  
\setlength{\tabcolsep}{\baselineColSep}
\par\noindent\makebox[\linewidth][c]{%
\begin{tabular}{lrrrrrrrrrrrrrrrr}
\multicolumn{1}{l}{} & \multicolumn{1}{l}{} & \multicolumn{1}{l}{} & \multicolumn{2}{c}{\makecell{WildlifeReID\\10K}} & \multicolumn{2}{c}{\makecell{iNat2021\\mini}} & \multicolumn{2}{c}{Omniglot} & \multicolumn{2}{c}{FungiTastic} & \multicolumn{6}{c}{\makecell{State Information Prediction\\NMSE$\downarrow \times$100}} \\
\cmidrule(lr){4-5} \cmidrule(lr){6-7} \cmidrule(lr){8-9} \cmidrule(lr){10-11} \cmidrule(lr){12-17}
\multicolumn{1}{l}{} & \multicolumn{1}{l}{} & \multicolumn{1}{l}{\multirow[b]{2}{*}[-\cmidruleheight]{\makecell{Pre-train\\GFLOPs}}} & \multicolumn{2}{c}{mAP} & \multicolumn{2}{c}{kNN} & \multicolumn{2}{c}{Acc.} & \multicolumn{2}{c}{kNN} & \multicolumn{2}{c}{Pong} & \multicolumn{2}{c}{Dino} & \multicolumn{2}{c}{Golf} \\
\cmidrule(lr){4-5} \cmidrule(lr){6-7} \cmidrule(lr){8-9} \cmidrule(lr){10-11} \cmidrule(lr){12-13} \cmidrule(lr){14-15} \cmidrule(lr){16-17}
Model & Ep. & \multicolumn{1}{l}{} & CLS & Pool & CLS & Pool & CLS & Pool & CLS & Pool & CLS & Pool & CLS & Pool & CLS & Pool \\
\midrule
Bootleg & 600 & 4.15 & 18.0 & 16.7 & 18.7 & 11.9 & 42.0 & 45.8 & 31.4 & 27.0 & 0.8 & 3.2 & 8.5 & 12.5 & 6.1 & 6.8 \\
LeJEPA & 100 & 2.01 & 19.8 & 18.8 & 13.7 & 9.5 & 43.3 & 46.5 & 29.2 & 24.4 & 2.6 & 1.4 & 5.7 & 5.2 & 12.5 & 10.2 \\
ColorMAE & 300 & 1.08 & 15.0 & 12.6 & 6.9 & 6.4 & 54.8 & 43.0 & 17.2 & 16.7 & 2.1 & 20.1 & 3.2 & 5.4 & 5.9 & 6.0 \\
ColorMAE & 800 & 2.87 & 14.3 & 11.8 & 7.6 & 7.3 & 48.3 & 40.0 & 17.3 & 17.5 & 1.1 & 2.1 & 10.5 & 3.3 & 3.8 & 8.0 \\
ColorMAE & 1600 & 5.74 & 14.7 & 11.7 & 9.4 & 7.9 & 51.0 & 45.0 & 18.7 & 18.2 & 13.0 & 5.2 & 3.9 & 4.4 & 4.3 & 19.2 \\
MAE & 1600 & 5.74 & 10.9 & 10.8 & 6.9 & 6.1 & 37.3 & 39.8 & 16.0 & 15.5 & 3.4 & 10.2 & 8.3 & 3.5 & 28.3 & 12.3 \\
SimDINOv2 & 100 & 2.91 & 23.2 & 19.8 & 33.9 & 14.6 & 66.0 & 66.3 & 39.7 & 27.8 & 1.7 & 0.6 & 5.3 & 4.4 & 7.5 & 5.9 \\
\bottomrule
\end{tabular}
}%
\par
\endgroup

\vspace*{\fill}
\clearpage

\begingroup
\tiny
\providecommand{\baselineColSep}{2pt}  
\setlength{\tabcolsep}{\baselineColSep}
\providecommand{\baselineVRule}{\hskip-\arrayrulewidth\vrule width\arrayrulewidth}
\par\noindent\makebox[\linewidth][c]{%
\begin{tabular}{l!{\baselineVRule}l!{\baselineVRule}l!{\baselineVRule}l!{\baselineVRule}rrrrrrrrrrrrrrrrrrrrr}
\multicolumn{1}{l}{} & \multicolumn{1}{l}{} & \multicolumn{1}{l}{} & \multicolumn{1}{l}{} & \multicolumn{5}{c}{IN-1K} & \multicolumn{2}{c}{\makecell{IN-21K\\10-shot}} & \multicolumn{2}{c}{\makecell{m-VTAB\\Natural}} & \multicolumn{2}{c}{\makecell{m-VTAB\\Specialized}} & \multicolumn{2}{c}{\makecell{m-VTAB\\Structured}} & \multicolumn{8}{c}{\makecell{Instance Retrieval\\mAP}} \\
\cmidrule(lr){5-9} \cmidrule(lr){10-11} \cmidrule(lr){12-13} \cmidrule(lr){14-15} \cmidrule(lr){16-17} \cmidrule(lr){18-25}
\multicolumn{1}{l}{} & \multicolumn{1}{l}{} & \multicolumn{1}{l}{\multirow[b]{2}{*}[-\cmidruleheight]{\makecell{Aug.\\Strength}}} & \multicolumn{1}{l}{\multirow[b]{2}{*}[-\cmidruleheight]{\makecell{Color\\Mask}}} & \multicolumn{2}{c}{kNN} & \multicolumn{2}{c}{LP} &  & \multicolumn{2}{c}{kNN} & \multicolumn{2}{c}{kNN} & \multicolumn{2}{c}{kNN} & \multicolumn{2}{c}{kNN} & \multicolumn{2}{c}{ROxford} & \multicolumn{2}{c}{RParis} & \multicolumn{2}{c}{ILIAS} & \multicolumn{2}{c}{SOP} \\
\cmidrule(lr){5-6} \cmidrule(lr){7-8} \cmidrule(lr){10-11} \cmidrule(lr){12-13} \cmidrule(lr){14-15} \cmidrule(lr){16-17} \cmidrule(lr){18-19} \cmidrule(lr){20-21} \cmidrule(lr){22-23} \cmidrule(lr){24-25}
\multicolumn{1}{l}{Swing} & \multicolumn{1}{l}{Ep.} & \multicolumn{1}{l}{} & \multicolumn{1}{l}{} & CLS & Pool & CLS & Pool & \multicolumn{1}{c}{FT} & CLS & Pool & CLS & Pool & CLS & Pool & CLS & Pool & CLS & Pool & CLS & Pool & CLS & Pool & CLS & Pool \\
\bottomrule
\end{tabular}
}%
\par
\endgroup

\clearpage
\begingroup
\tiny
\providecommand{\baselineColSep}{2pt}  
\setlength{\tabcolsep}{\baselineColSep}
\providecommand{\baselineVRule}{\hskip-\arrayrulewidth\vrule width\arrayrulewidth}
\par\noindent\makebox[\linewidth][c]{%
\begin{tabular}{l!{\baselineVRule}l!{\baselineVRule}l!{\baselineVRule}l!{\baselineVRule}rrrrrrrrrrrrrr}
\multicolumn{1}{l}{} & \multicolumn{1}{l}{} & \multicolumn{1}{l}{} & \multicolumn{1}{l}{} & \multicolumn{2}{c}{\makecell{WildlifeReID\\10K}} & \multicolumn{2}{c}{\makecell{iNat2021\\mini}} & \multicolumn{2}{c}{Omniglot} & \multicolumn{2}{c}{FungiTastic} & \multicolumn{6}{c}{\makecell{State Information Prediction\\NMSE$\times$100$\downarrow$}} \\
\cmidrule(lr){5-6} \cmidrule(lr){7-8} \cmidrule(lr){9-10} \cmidrule(lr){11-12} \cmidrule(lr){13-18}
\multicolumn{1}{l}{} & \multicolumn{1}{l}{} & \multicolumn{1}{l}{\multirow[b]{2}{*}[-\cmidruleheight]{\makecell{Aug.\\Strength}}} & \multicolumn{1}{l}{\multirow[b]{2}{*}[-\cmidruleheight]{\makecell{Color\\Mask}}} & \multicolumn{2}{c}{mAP} & \multicolumn{2}{c}{kNN} & \multicolumn{2}{c}{Acc.} & \multicolumn{2}{c}{kNN} & \multicolumn{2}{c}{Pong} & \multicolumn{2}{c}{Dino} & \multicolumn{2}{c}{Golf} \\
\cmidrule(lr){5-6} \cmidrule(lr){7-8} \cmidrule(lr){9-10} \cmidrule(lr){11-12} \cmidrule(lr){13-14} \cmidrule(lr){15-16} \cmidrule(lr){17-18}
\multicolumn{1}{l}{Swing} & \multicolumn{1}{l}{Ep.} & \multicolumn{1}{l}{} & \multicolumn{1}{l}{} & CLS & Pool & CLS & Pool & CLS & Pool & CLS & Pool & CLS & Pool & CLS & Pool & CLS & Pool \\
\midrule
\multirow{27}{*}{None} & \multirow{9}{*}{100} & \multirow{3}{*}{Weak} & None & \cellcolor{bad_color!20}9.4 & \cellcolor{bad_color!18}12.5 & \cellcolor{bad_color!20}0.8 & \cellcolor{bad_color!20}3.0 & \cellcolor{bad_color!20}11.3 & \cellcolor{bad_color!20}21.0 & \cellcolor{bad_color!20}5.8 & \cellcolor{bad_color!20}10.0 & \cellcolor{good_color!7}2.6 & \cellcolor{good_color!13}1.2 & \cellcolor{bad_color!6}9.2 & \cellcolor{good_color!4}4.5 & \cellcolor{bad_color!1}14.1 & \cellcolor{bad_color!2}15.2 \\
\bottomrule
\end{tabular}
}%
\par
\endgroup

\end{document}